\documentclass{article}

\usepackage[preprint, nonatbib]{neurips_2020}

\usepackage[utf8]{inputenc} % allow utf-8 input
\usepackage[T1]{fontenc}    % use 8-bit T1 fonts
\usepackage{hyperref}       % hyperlinks
\usepackage{url}            % simple URL typesetting
\usepackage{booktabs,multirow}       % professional-quality tables
\usepackage{amsfonts}       % blackboard math symbols
\usepackage{nicefrac}       % compact symbols for 1/2, etc.
\usepackage{microtype}      % microtypography

\usepackage{graphicx}
\usepackage{times}
\usepackage{epsfig}
\usepackage{graphicx}
\usepackage{amsmath}
\usepackage{amssymb}
\usepackage{xcolor,vwcol}

\DeclareMathAlphabet{\pazocal}{OMS}{zplm}{m}{n} %imkl

\usepackage{mathtools}
\DeclarePairedDelimiterX{\norm}[1]{\lVert}{\rVert}{#1}

\usepackage{multicol}

\usepackage[linesnumbered,ruled,vlined]{algorithm2e}

\usepackage{setspace}

\let\oldnl\nl% Store \nl in \oldnl
\newcommand{\nonl}{\renewcommand{\nl}{\let\nl\oldnl}}% Remove line number for one line

\title{MTAdam: Automatic Balancing of\\ Multiple Training Loss Terms}

\author{Itzik Malkiel \\
  Tel Aviv University \\
\\\And
  Lior Wolf \\
  Tel Aviv University \\
  Facebook AI Research \\
}

\begin{document}

\maketitle

\begin{abstract}
When training neural models, it is common to combine multiple loss terms. The balancing of these terms requires considerable human effort and is computationally demanding. Moreover, the optimal trade-off between the loss term can change as training progresses, especially for adversarial terms. In this work, we generalize the Adam optimization algorithm to handle multiple loss terms. The guiding principle is that for every layer, the gradient magnitude of the terms should be balanced. To this end, the Multi-Term Adam (MTAdam) computes the derivative of each loss term separately, infers the first and second moments per parameter and loss term, and calculates a first moment for the magnitude per layer of the gradients arising from each loss. This magnitude is used to continuously balance the gradients across all layers, in a manner that both varies from one layer to the next and dynamically changes over time. Our results show that training with the new method leads to fast recovery from suboptimal initial loss weighting and to training outcomes that match conventional training with the prescribed hyperparameters of each method. MTAdam code can be found here\footnote{https://github.com/ItzikMalkiel/MTAdam}.
\end{abstract}

\section{Introduction}

In both supervised and unsupervised image generation tasks, one tries to balance a set of criteria, some overlapping and some presenting natural tradeoffs. For example, in image analogy tasks, multiple conditioned generation criteria, some based on comparing the input to the output using an MSE term and some on perceptual losses, may be combined with one or more adversarial losses. If the mapping is two-sided, one also employs a circularity loss. If the image is of a composite nature, mask-based losses are added, and so on.

It has been demonstrated repeatedly that adding losses can be beneficial. In many cases, formulating the desiderata as loss terms is well paved. However, as more loss terms are added, the space of possible balancing weights increases exponentially, and more resources need to be allocated to identify good configurations that would justify the added terms. Another common challenge is that at different stages of the training process, the optimal balance may change~\cite{mescheder2017numerics}. This is especially true for adversarial losses, which often become unstable. It is, therefore, necessary to have the balancing terms update dynamically during training.

In this work, we introduce Multi-Term Adam (MTAdam), an optimization algorithm for multi-term loss functions. MTAdam extends Adam~\cite{kingma2014Adam} and allows an effective training of an unweighted multi-term loss objective. Thus, MTAdam can streamline the computationally demanding task of hyperparameter search, required for effectively weighting multi-term loss objectives.

At every training iteration, a dynamic weight is assigned to each of the loss terms based on the magnitude of the gradient that each term entails. The weights are assigned in a way that balances between these gradients and equates their magnitude. 

This, however, would be an ineffective balancing method, without three crucial components: (i) the balancing needs to occur independently for each layer of the neural network, since the relative contributions of the losses vary, depending on the layer, (ii) the balancing needs to be anchored by a dominant loss term in order to allow a natural progression of the effective learning rate, i.e., one cannot normalize such that the magnitude becomes a constant, and (iii) the update step needs to take into account the maximal variance among all losses, to support sufficient explorations in places of the parameter space, in which one of the losses becomes more sensitive. 

The main focus of our experiments is in the field of conditional adversarial image generation. This domain is known to require multiple loss terms. Moreover, the quality of the generated images is evaluated using acceptable success metrics that are not directly optimized by any of the loss terms, such as FID~\cite{fid}. Our results show that MTAdam is able to recover from unbalanced starting points, in which the weight parameters are set inappropriately, while Adam and other baseline methods cannot.

\section{Related Work}

SGD with momentum (Nesterov)~\cite{rumelhart1986learning} is an optimization algorithm that extends SGD, suggesting to update the network's parameters by a moving average of the gradients, rather than the gradients at each step. Root Mean Square Propagation (RMSProp)~\cite{rmsprop}  extends SGD by dividing the learning rate during the backward step, by the moving average of the second moment of the gradients of each parameter. Adam~\cite{kingma2014Adam} combines both principles and employs the first and second moments of the gradients of each learned parameter and applies them during the backward step. Adam has become a dominant optimizer, that is applied across many applications, and, in particular, it is the de-facto standard in the field of adversarial training. It is known for improving convergence to work well with the default values of its own hyperparameters ($\beta_1$ and $\beta_2$). 

Multiple methods have been suggested for selecting hyperparameters~\cite{srinivas2009gaussian, wang2016optimization, bergstra2011algorithms, feurer2014using}.  Hyperband~\cite{li2017hyperband} performs hyperparameter search as an infinite-armed bandit problem utilizing a predefined amount of resources while searching for the best configuration of hyperparameters that maximizes the given success criterion. It can provide an order-of-magnitude speedup compared to Bayesian optimization~\cite{bergstra2011algorithms, snoek2012practical, hutter2011sequential}. We utilize Hyperband as a baseline that can directly optimize the FID score.

Image generation techniques utilize multiple loss terms, including pixel-wise norms, perceptual losses~\cite{johnson2016perceptual}, and one or more adversarial loss terms~\cite{NIPS2014_5423}. The need for a careful configuration of the weight parameters hinders research in this field and increases its costs.

In our experiments, we employ three different image generation methods, which vary in the type of supervision  employed or the task being solved: (i) pix2pix~\cite{isola2017image} generates an image in domain $B$ based on an input image in domain $A$, after observing matching pairs during training. (ii) CyclgeGAN~\cite{zhu2017unpaired} performs the same task while training in an unsupervised manner on unmatched images from the two domains. (iii) SRGAN~\cite{ledig2017photo} generates high-resolution images from low-resolution ones and is trained in a supervised way.

Multi-task learning shares some similarities with multi-term learning. In multi-task learning, one learns a few tasks concurrently, each associated with one or more loss terms. The most common approaches utilize hard parameter sharing~\cite{caruana1997multitask, kendall2018multi, he2017mask, long2015learning, ren2015faster}, in which a single model, with multiple task-specific heads, is trained. This approach is extremely effective when the tasks are highly related, as in MaskRCNN~\cite{he2017mask}. In the soft parameter sharing approach, each task has its own model, and a regularization term encourages the parameters of the different models to be similar~\cite{yang2016trace, duong2015low}. While MTAdam can be applied to multi-task learning (of both types) without substantial modifications, we leave the empirical validation of this ability for future investigations.

\section{Method}

The Adam algorithm optimized one stochastic objective function $f_t(\theta)$ over the set of parameters $\theta$, where $t$ is an index of the current mini-batch of samples. In contrast, MTAdam optimizes a set of such terms $ f_t^1(\theta),\dots,f_t^I(\theta)$. While Adam's task is to minimize the expected value $E_t[f_t(\theta)]$ w.r.t. the parameters $\theta$, MTAdam minimizes a weighted average of the $I$ terms. The weights of these mixtures are all positive, but otherwise unknown. The guiding principle for the determination of the weights at each iteration $t$ is that the moving average of the magnitude of the gradient of each term is equal across terms. This magnitude is evaluated and balanced at every layer of the neural network.

In Adam, two moments are continuously updated, using a moving average scheme: $m_t$ is the first moment of the gradient $\nabla_{\theta} f_t$ and $v_t$ is the second moment. Both are vectors of the same size of $\theta$. The moving averages are computed using the mixing coefficients $\beta_1$ and $\beta_2$ for the two moments.

MTAdam records such moments for each term $i=1...I$ separately. In addition, it uses a mixing coefficient $\beta_3$ in order to maintain the moving average of the gradient magnitude per each layer $\ell$, which is denoted by $n_{l,t}^i$.

Adam borrows from the SGD with momentum method (Nesterov) and updates the vector of parameters based on the weighted first moment of the gradient. In MTAdam, the first moment is computed based on a weighted gradient, in which the parameters of each layer $\ell$ for every term $i$ are weighted such that their magnitude is normalized by the factor $n_{\ell,t}^i$.  This way, across all layers, and at every time point, the $I$ terms contribute equally to the gradient step.

The Adam optimization algorithm is depicted in the left side of Alg.~\ref{algo:MTAdam} and MTAdam on the right. In line 1, MTAdam algorithm initializes $I$ pairs of first and second moment vectors. This is similar to Adam, except for initializing a pair of moments for each loss term. In line 2, and different from Adam, MTAdam initializes $I$ first moments for the magnitude of the gradients, per layer. In line 3, both MTAdam and Adam iterate over the stochastic mini-batches, performing $T$ training steps. 

In line 4, MTAdam iterates over the loss terms. For each loss term, MTAdam calculates its gradients over each one of the network layers (line 5), in an analogous manner to the way Adam computes the (single) gradients vector $\nabla_\theta f_t(\theta_t-1)$. In lines 6-8, MTAdam iterates over the layers, updates the moving average of the magnitude for each layer and loss term, and normalizes the gradients of the current layer and loss term, by multiplying with $\frac{n_{\ell,t}^1}{n_{\ell,t}^i}$. This multiplication normalizes the magnitude of the current gradients of layer $\ell$ and loss term $i$ using the moving average $n_\ell^i$. This normalization leads to all gradient magnitudes to be similar to that of the first loss term.

In line 5, we calculate the gradients of each specific loss term $i$ w.r.t $\theta$, across all layers. For a layer index $\ell$, the gradient is denoted by $g_\ell^i$. We denoted by $g^i$ the concatenation of all per-layer gradients.

Lines 6-8 of MTAdam do not have an Adam analog. The operation performed normalizes the magnitude of the gradients of all loss terms, to match the magnitude of the first loss term. This assigns a unique role to the first term, as the primary loss to which all other losses are compared. By linking the magnitude to that of a concrete loss, and not to a static value (e.g., normalizing to have a unit norm), we maintain the relationship between the training progression and the learning rate. 

The normalization iterates over the loss terms, and for each gradient, the first moment of the magnitude of the gradient is updated. The gradient magnitude is then normalized by that of the first loss term. 

Then, in lines 9-12, MTAdam updates the first and second moments for each parameter and each loss term and computes their bias correction. This is similar to Adam, except that the moments are calculated separately for each loss term. In lines 13-15, MTAdam iterates over the loss terms and calculates the steps from each term. The steps are summed over $\theta_{t-1}$, and the result is assigned $\theta_t$. 

In Adam, the update size is normalized by the second moment. In MTAdam, we divided by the maximal second moment among all loss terms. This division allows MTAdam to make smaller gradient steps, when a lower certainty is introduced by at least one of the loss terms. The motivation for this is that even if one of the losses is in a high-sensitivity region, where small updates create rapid changes to this term, then the step, regardless of the term which led to it, should be small. The importance of this maximization is demonstrated in the ablation study in Sec.~\ref{ablation}.

\begin{algorithm}[t]% [t]
\caption{Adam (left) and Multi-term Adam (right). All operations are element-wise. }
\label{algo:MTAdam}
\setlength{\columnsep}{-1.6cm}
\begin{multicols}{2}
% \begin{algorithmic}
% \vspace{-0.2cm}
%\scriptsize
\small
\DontPrintSemicolon
\KwIn{$\alpha$: step size, 
$\{\beta_1, \beta_2\}$: decay rates to\\ calculate the first and second moments,  $\theta_0$:\\ initial weights, $f_t(\theta)$: stochastic objective. \\
}
\KwOut{$\theta_t$: resulting parameters}
$m_0, v_0 \gets 0, 0$ %(Initialize moving 1st and 2nd moment)
\setstretch{1.15}
\;
\;
\While{$t = \{1, \cdots, T\}$}{
    \;
    $g \gets \nabla_{\theta} f_t(\theta_{t - 1})$\;

    \;
    \;
    % \nonl  \;
    \;
    $m_{t} \leftarrow \beta_{1} \cdot m_{t-1}+\left(1-\beta_{1}\right) \cdot g$  \;
    
    $v_{t} \leftarrow \beta_{2} \cdot v_{t-1}+\left(1-\beta_{2}\right) \cdot {g^{2}}$  \;
    
    $\widehat{m_t} \leftarrow m_{t} /\left(1-\beta_{1}^{t}\right)$  \;
    
    $\widehat{v_t} \leftarrow v_{t} /\left(1-\beta_{2}^{t}\right)$ \;
    %  \nonl  \;
    \vspace{+0.2cm}
    \;
    \vspace{+0.1cm}
    $\theta_{t} \leftarrow \theta_{t-1}-\alpha \cdot \widehat{m_t} /(\sqrt{\widehat{v_t}}+\epsilon)$\;
    \vspace{+0.1cm}
    \;
}
\Return{$\theta_T$}

\columnbreak
\setcounter{AlgoLine}{0}
\setstretch{1}

\small
\DontPrintSemicolon
\KwIn{In addition to Adam's parameters: $\beta_3$: decay rate for the gradient's norm first moment,
%$\theta_0$: initial parameter, $f^1_t(\theta)$...$f^I_t(\theta)$ and $r_1...r_I$: stochastic multiple objective functions and their ranking, respectively}
$f^1_t(\theta)$...$f^I_t(\theta)$: stochastic loss term functions.}
\KwOut{$\theta_t$: resulting parameters}
\lFor{$i = \{1, \cdots, I\}$}{    $m^i_0, v^i_0 \gets 0, 0$ %(Initialize moments estimations) %(Initialize 1st and 2nd moment vectors, for each loss term)\; % n^i_0is equivalent to p.norms
}
\lFor{$l$ in $1...L$}{
$n_{\ell,0}^i \gets 1$ % (Initialize magnitude moments)%(Initialize 1st moment scalar for the enregy of each loss terms, per layer. )\; % n^i_0is equivalent to p.norms
}
% $\rho_\infty \gets 2/(1 - \beta_2) - 1$ (Compute the maximum length of the approximated SMA)\;
\While{$t = \{1, \cdots, T\}$}{
    % $g_t \gets \Delta_{\theta} f_t(\theta_{t - 1})$ (Calculate gradients w.r.t. stochastic objective at timestep t)\;
    \For{$i = \{1, \cdots, I\}$}{
        {$g^i$ := $(g^i_1, ... g^i_L) $ }$\gets \nabla_{\theta} f^i_t(\theta_{t - 1})$ % (Calculate gradients w.r.t. the $i^{th}$ stochastic objective at timestep t)\;

        \For{$\ell$ in $1...L$}{
        
            $n_{\ell,t}^i \leftarrow \beta_{3} \cdot n_{\ell,t-1}^i+\left(1-\beta_{3}\right) \cdot {\norm{g^i_\ell}}_2$             %($g^i_\ell$ is the $g^i$ gradients on $l$ )% (Update biased gradient norms estimation, per layer.) \;
        
            %$\widehat{({n_l^i})_t} \leftarrow ({n_l^i})_t /\left(1-\beta_{3}^{t}\right)$ (Compute bias-corrected gradient norm estimate)\;
            % grad = p.norms[anchor_index] * grad / p.norms[loss_index] #  p.norms[0] is the anchor
            
            $g^i_\ell \gets n_{\ell,t}^1 \cdot g^i_\ell / (n_{\ell,t}^i)$ % (normalizing the gradient norms, per layer, to the layer's anchor norm. )% assuming the first loss is the anchor - for which all other loss terms will get align with. In multi term 
        }
        
        $m^i_{t} \leftarrow \beta_{1} \cdot m^i_{t-1}+\left(1-\beta_{1}\right) \cdot g^i$ %(Update biased first moment estimate) 
        \;
        
        $v^i_{t} \leftarrow \beta_{2} \cdot v^i_{t-1}+\left(1-\beta_{2}\right) \cdot {(g^i)}^{2}$ %(Update biased second raw moment estimate) 
        \;

        $\widehat{m^i_t} \leftarrow m^i_{t} /\left(1-\beta_{1}^{t}\right)$ %(Compute bias-corrected first moment estimate) 
        \;
        
        $\widehat{v^i_t} \leftarrow v^i_{t} /\left(1-\beta_{2}^{t}\right)$ %(Compute bias-corrected second raw moment estimate)
        \;
        
    }
    
    \For{$i = \{1, \cdots, I\}$}{
            $\theta_{t-1} \leftarrow \theta_{t-1}-\alpha \cdot \widehat{m^i_t} /(\sqrt{\max(\widehat{v^1_t}...\widehat{v^I_t}})+\epsilon)$ 
    }
    $\theta_{t} \gets \theta_{t-1}$
}
\Return{$\theta_T$}
\end{multicols}
\end{algorithm}

\noindent{\bf Memory and Run Time Analysis\quad}
Adam utilizes a pair of 1st and 2nd moments for each learned parameter. Given a network with $\norm{\theta}$ learned parameters, it has a memory complexity of $O({|\theta|})$. MTAdam utilizes $I$ different pairs of 1st and 2nd moments for each parameter. In addition, MTAdam employs $I$ first moments magnitude for each layer.  These two extensions bring the memory complexity to $O(I {|\theta|} + IL)\sim O(I {|\theta|})$. In MTAdam, the run time complexity also depends on the number of loss terms  $O(I \theta )$. The dependence on the number of layers $L$ in Alg.~\ref{algo:MTAdam} can be absorbed in $\Theta$.

\section{Experiments}
\label{results}

We compare the results of MTAdam with five baselines: (1-3) Adam, RMSProp, and SGD with momentum, applied with unbalanced weightings. (4) Hyperband~\cite{li2017hyperband} applied to perform a hyper-parameter search for the lambdas between the loss terms, utilizing the FID metric. (5) Adam optimizer applied with a balanced weighting. 

We note that baseline (4) benefits from running training multiple times and that baseline (5) employs the weights proposed for each method after a development process that is likely to have included a hyper-parameter search, in which multiple runs were evaluated by the developers of each method. For each optimization method, we employ the default parameters in pytorch. For the Adam experiments, we set the hyperparameters $\beta_1$ and $\beta_2$ to 0.9 and 0.999, respectively. For MTAdam, $\beta_1$ and $\beta_2$ are configured with the same values as Adam and $\beta_3=0.9$.

\subsection{MNIST classification}

In order to turn MNIST to an unbalanced multi-term experiment, we compute the loss for each of the digits separately, creating ten loss terms, each weighted by a random weight from the uniform distribution between 1 and 1000. The test set is unweighted, which causes classes that are associated with lower weights to suffer from underfitting.

The official of the pytorch MNIST example is used: two convolutional layers followed by two fully connoted layers. The experiment is repeated 100 times, and for the sake of saving computations, hyperband is not tested. The results in Tab.~\ref{tab:mnist}, show a clear advantage for MTAdam over the other unbalanced alternatives.

\begin{table*}[t!]
\caption{MNIST results. Shown is mean accuracy in percents and SD for 100 runs. }
\label{tab:mnist}
\centering
\begin{tabular}{ccccc}
\toprule
\multicolumn{4}{c}{Unbalanced} & Balanced\\
\cmidrule{1-4}
%\cmidrule(lr){5}
SGD-Momentum & RMSProp & Adam & MTAdam & Adam\\
\midrule

85.4 $\pm$ 1.23  &87.2 $\pm$ 0.97  &88.8 $\pm$ 0.84  & 97.9 %96.5
$\pm$ 0.07 & 98.3 $\pm$ 0.05 \\
\bottomrule
\end{tabular}
\end{table*}

\subsection{Image synthesis}
We demonstrate the ability of MTAdam to effectively converge when applied with unbalanced multi-term loss objectives. To this end, we compare the performance of MTAdam with other optimizers, evaluated on three methods, pix2pix~\cite{isola2017image}, CycleGan~\cite{zhu2017unpaired} and SRGAN~\cite{ledig2017photo}. 

We used the learning rate as found in the implementation of each method~\cite{isola2017image,zhu2017unpaired,ledig2017photo}. Performance is evaluated using various metrics: L1, L2, PSNR, NMSE (normalized MSE), FID~\cite{fid}, and SSIM~\cite{wang2004image}. In each case, following the metrics used in the original work, with the addition of FID.

\begin{figure*}[t]
\centering
\begin{tabular}{cc}
    \includegraphics[width=0.45\linewidth]{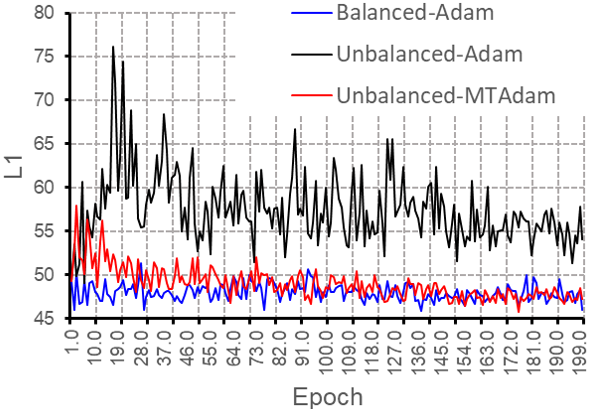}&
    \includegraphics[width=0.45\linewidth]{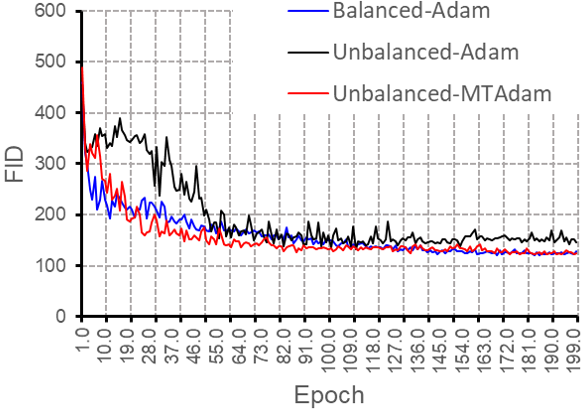}\\
    \end{tabular}
  \centering
  \caption{(left) L1 and (right) FID per epoch on the validation set of Facade for three pix2pix variants. Balanced-Adam and Unbalanced-Adam employ Adam, with a balanced and unbalanced weighting, respectively. Unbalanced-MTAdam utilizes MTAdam.
  }
    \label{fig:pix2pix_per_epoch}
\end{figure*}
\begin{figure*}[t]
  \includegraphics[width=0.9\linewidth]{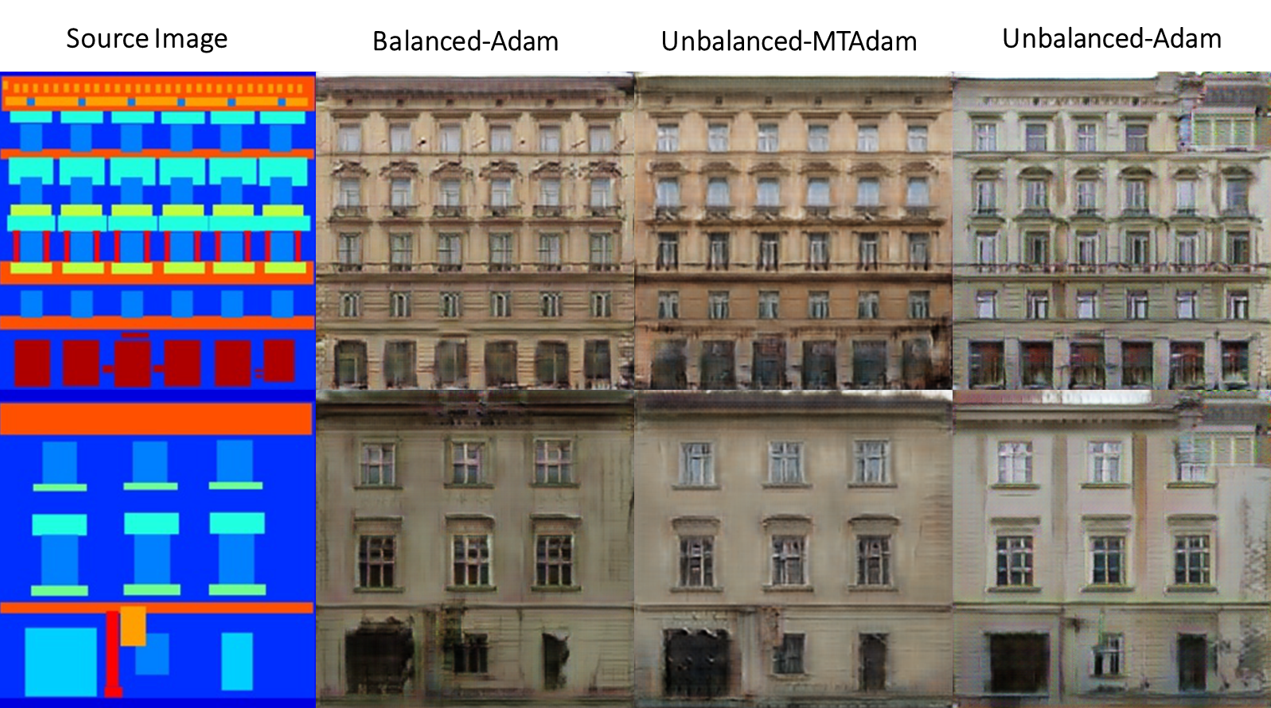}
  \centering
  \caption{Samples from the Facade test set. Balanced-Adam employs pix2pix with the prescribed weights~\cite{isola2017image}. Unbalanced-Adam runs with equal initial weights, which leads to visual artifacts.  Unbalanced-MTAdam (our method) starts with equal weights, but produces images of the same quality as Balanced-Adam.}
    \label{fig:quali_pix2pix}
\end{figure*}

{\bf Pix2pix Experiments\quad} The objective function of the pix2pix generator has dual-terms:
\begin{equation}
\pazocal{L}_{pix2pix} = \lambda_1 \pazocal{L}_{1} + \lambda_2 \pazocal{L}_{GAN}
\end{equation}
Where $\pazocal{L}_{GAN}$ is the GAN loss of the generator, $\pazocal{L}_{1}$ is the pixel loss, and $\lambda_1$ and $\lambda_2$ are set to 100 and 1, respectively. In our study, we unbalance pix2pix models by setting $\lambda_2$ to 100 (which implies a 1:1 ratio between the two loss terms). 

Two datasets are used: matching facade images and their semantic labels~\cite{tylevcek2013spatial} and aerial photographs and matching maps~\cite{isola2017image}. Performance is reported on a holdout test set of each benchmark.

Fig.~\ref{fig:pix2pix_per_epoch}, depicts the test-performance of multiple models per each training epoch. The experiment's name contains `unbalanced' for the case of equal loss terms, and `balanced' when using the prescribed $\lambda$ values. As can be seen, unbalanced-MTAdam yields a similar convergence as the balanced-Adam. Specifically, the MTAdam experiment converges substantially better than the unbalanced Adam experiment,  leading to improved L1 and FID scores. 

In Tab.~\ref{Tab:pix2pix_test} MTAdam is compared with all baselines, applied for training pix2pix models. The top four models, in each section, utilize an identical unbalanced setting, each applied with a different optimizer. The Hyperband experiment utilizes the FID metric to perform a hyperparameter search on $\lambda_2$. The Hyperband experiments incorporate 40 trials (i.e. 40 training processes, which gives it a great advantage), each trial randomly sampled a different lambda from the range $[10^{-4},10^4]$. We report Hyperband performance by utilizing the trial that is associated with the chosen lambda. In the aerial photograph experiment, the Hyperband failed to choose a $\lambda_2$ value that is close to the original value of 1. In the facade experiment, Hyperband sampled at least one lambda value between 0.5 to 10, yet the retrieved best model utilizes a higher lambda value of 162.89, since this value showed a preferable FID value on the validation set. In the maps dataset, a value of 59.61 was selected.

The results of the table clearly show the advantage of MTAdam over all baseline methods. In addition, it also shows a slight improvement in performance in comparison to the usage of Adam on the prescribed weights. Fig.~\ref{fig:quali_pix2pix} presents two representative samples from the facades test set. Pix2pix-Unbalanced introduces visual artifacts and suffers from mode collapse (it generates the same corrupted patch in the top right corner of many images). Pix2pix-Unbalanced-MTAdam yields higher-quality images, similar to those of the original Pix2Pix, using the prescribed weights.

\begin{table*}[t!]
\caption{Comparing various methods for traning pix2pix. Lower is better for all metrics. }
\label{Tab:pix2pix_test}
\centering
\resizebox{1\linewidth}{!}{
\begin{tabular}{lc@{~~}c@{~~}c@{~~}cc@{~~}c@{~~}c@{~~}c}
\toprule
&\multicolumn{4}{c}{Facade images} & \multicolumn{4}{c}{maps$\rightarrow$aerial}\\
\cmidrule(lr){2-5}\cmidrule(lr){6-9}
&  \textbf{L1} &  \textbf{L2} & \textbf{NMSE} & \textbf{FID}& \textbf{L1} &  \textbf{L2} & \textbf{NMSE} & \textbf{FID}  \\
\midrule
Unbalanced-SGD-Momentum &120.12&  18308 & 1.132  &  339.40 & 45.1 & 3428.3 & 0.491  &  102.45 \\
Unbalanced-RMSProp & 53.98 &  4913 & 0.303  &  136.10& 42.4 & 3056.7 & 0.435  &  95.74  \\
Unbalanced-Adam &54.92&  5074& 0.313  &  133.55 & 38.1 & 2632.3 & 0.368  & 88.08 \\

Unbalanced-MTAdam & 46.50 & 3822 & 0.236 & \textbf{130.38}& {\textbf{33.6}} & {\textbf{2096.9}} & {\textbf{0.293}} & 86.40\\
\midrule
Hyperband-FID  &54.12 & 4992 & 0.308 & 141.96 & 35.14 & 2391.2 & 0.341 & 86.45\\
\midrule
Balanced-Adam & {\textbf{45.84}} & {\textbf{3811}} & {\textbf{0.235}} &  130.50& 34.4 & 2163.4 & 0.303 &  {\textbf{85.76}} \\
\bottomrule
\end{tabular}}
\end{table*}

\paragraph{CycleGAN experiments} The CycleGAN objective function is composed of six loss terms:
\begin{multline}
\pazocal{L}_{CycleGAN} = \lambda_1 \pazocal{L}_{GAN\_A} + \lambda_1 \pazocal{L}_{GAN\_B} + \lambda_2 \pazocal{L}_{cycle\_A} + \lambda_2 \pazocal{L}_{cycle\_B} + \lambda_3 \pazocal{L}_{idt\_A} + \lambda_3 \pazocal{L}_{idt\_B}
\end{multline}
Where the $\pazocal{L}_{GAN}$, $\pazocal{L}_{cycle}$, $\pazocal{L}_{idt}$ terms are the GAN loss, cycle GAN loss and identity loss, for each one of the sides (A or B). $\lambda_1$, $\lambda_2$ and $\lambda_3$ are set to 1, 10 and 0.5, respectively. In our experiments, we unbalance CycleGAN, by setting $\lambda_2$ to 1000, leaving $\lambda_1$ and $\lambda_3$ unchanged. 

\begin{figure*}[t]

\begin{tabular}{cc}
    \includegraphics[width=0.45\linewidth]{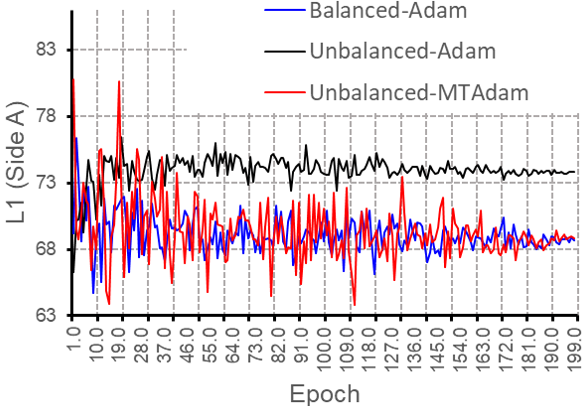}
    \includegraphics[width=0.45\linewidth]{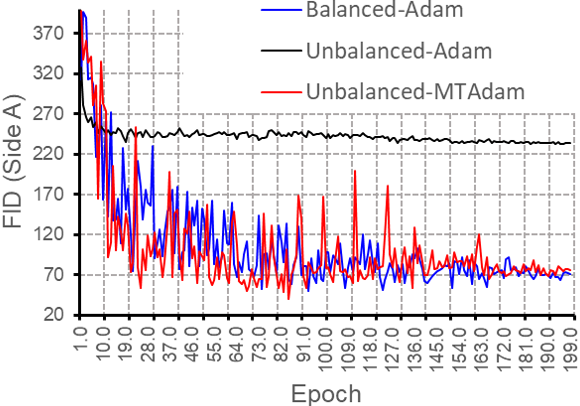} \\
    ~\\
    \includegraphics[width=0.45\linewidth]{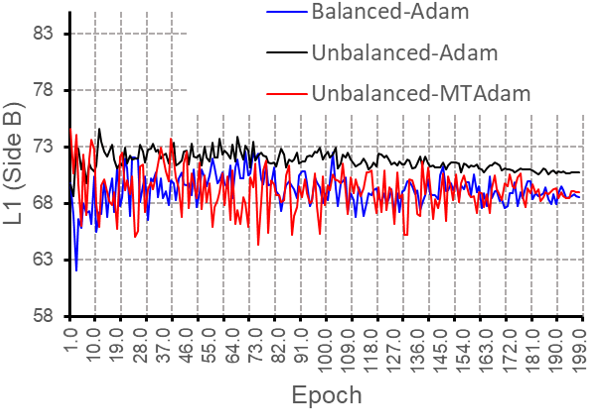}
    \includegraphics[width=0.45\linewidth]{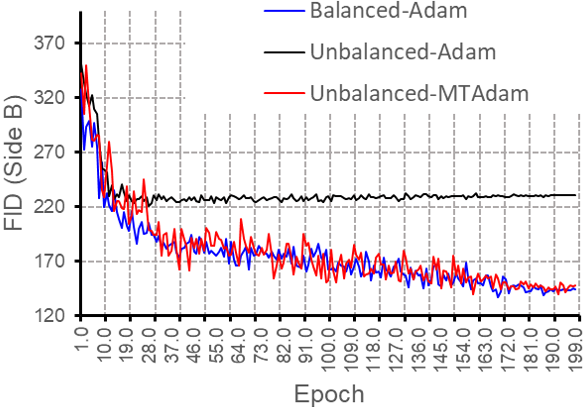} 
    \end{tabular}
  \centering
  \caption{The L1 (left) and FID (right) metrics evaluated per epoch CycleGAN models trained on the horse2zebra dataset~\cite{deng2009imagenet}. (top) generated zebra images. (bottom) generated horse images.
  }
    \label{fig:cyclegan_per_epoch}
\end{figure*}

Fig.~\ref{fig:cyclegan_per_epoch} compares the convergence of three CycleGAN models: unbalanced MTAdam, unbalanced Adam, and balanced Adam. All models are applied on the horse2zebra dataset~\cite{deng2009imagenet}. As can be seen, MTAdam exhibits a competitive convergence to the Adam experiment applied with balanced weighting, which is much better than the performance of Adam on the unbalanced weights. Tab.~\ref{Tab:cycleGAN_TEST} presents the performance of MTAdam applied on CycleGAN, compared to all five baselines, and evaluated on two datasets. MTAdam with an unbalanced initialization yields competitive performance to the Adam method, which uses the prescribed hyperparams. 

\begin{table*}[t]
\caption{Performance of the various optimization methods, when applied to the CycleGAN algorithm.}
\label{Tab:cycleGAN_TEST}
\centering
\resizebox{1\linewidth}{!}
{%
\begin{tabular}{ll@{~~}c@{~~}c@{~~}c@{~~}c@{~~}c@{~~}c@{~~}c@{~~}c}
\toprule
 &  & \textbf{L1-A} &  \textbf{L2-A} & \textbf{NMSE-A} & \textbf{FID-A} & \textbf{L1-B} &  \textbf{L2-B} & \textbf{NMSE-B} & \textbf{FID-B}\\
\midrule

\parbox[t]{2mm}{\multirow{6}{*}{\rotatebox[origin=c]{90}{\textbf{horse$\rightarrow$zebra}}}} & Unbalanced-SGD-Momentum & 82.3 & 10217 & 0.5866 & 283.3  & 127.0& 20331 & 0.9898 & 406.3\\
&Unbalanced-RMSProp & 73.8 & 8384 & 0.4814 &  233.9 & 70.6 &  7683 & 0.3740 & 230.9 \\

&Unbalanced-Adam & 74.4 & 8421 & 0.4935 & 229.6 & 72.3 & 7742   & 0.3842 & 227.1\\

&Unbalanced-MTAdam & 68.7 & \textbf{7235} & \textbf{0.4151} & 73.0 &   69.0&  7281 & 0.3545 & \textbf{139.7}\\
\cmidrule{2-10}
&Hyperband-FID
& 72.3 & 8049 & 0.4687 & 189.2  & 71.8 & 7584  &  0.3746 & 192.6 \\
\cmidrule{2-10}

&Balanced-Adam & \textbf{68.6} & 7248 & 0.4161 & \textbf{70.6} & \textbf{68.6} & \textbf{7207}  & \textbf{0.3509} & 143.8 \\
\midrule
\parbox[t]{2mm}{\multirow{6}{*}{\rotatebox[origin=c]{90}{\textbf{maps$\rightarrow$aerial}}}}&Unbalanced-SGD-Momentum &26.7  & 1102 &0.0209&437.5&73.01&7146&1.0&738.6\\
&Unbalanced-RMSProp & 12.9 &389&0.0074&170.2&36.81&2260&0.3154&231.3\\

&Unbalanced-Adam & 12.9 & 386 & 0.0073  & 166.3 & 37.00 & 2270 &  0.3161 & 234.2 \\

&Unbalanced-MTAdam & {10.6} & {376}  & {0.0071} & {\textbf{113.0}}  & {\textbf{35.37}}  & {\textbf{2242}} & {\textbf{0.3137}} & {48.8} \\
\cmidrule{2-10}
&Hyperband-FID
& 12.9 & 387 & 0.0073 & 169.4 & 36.91 & 2226 & 0.3102 & 233.1\\
\cmidrule{2-10}

&Balanced-Adam & \textbf{8.5} & \textbf{301} & \textbf{0.0057} & 116.1 & 35.45 & 2250 & 0.3148  & \textbf{46.8}\\

\bottomrule
\end{tabular}}
\end{table*}

\begin{figure*}[t]
\centering
\begin{tabular}{ccccc}
    \includegraphics[width=.168\linewidth]{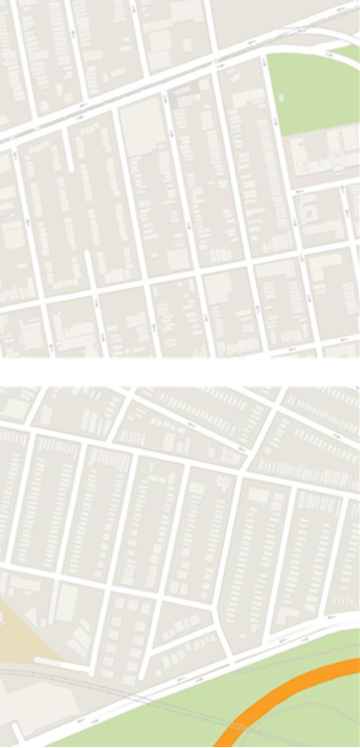}&
    \includegraphics[width=.168\linewidth]{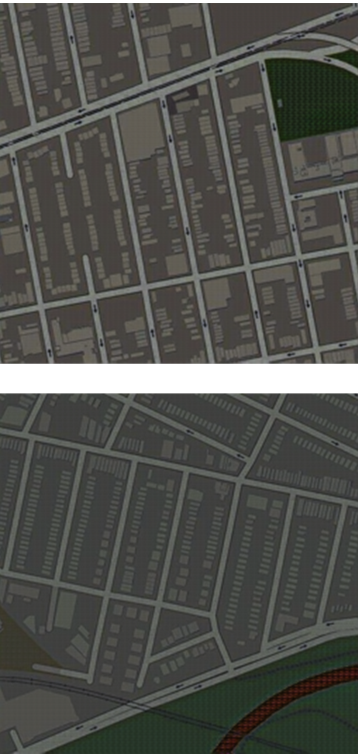}&
    \includegraphics[width=.168\linewidth]{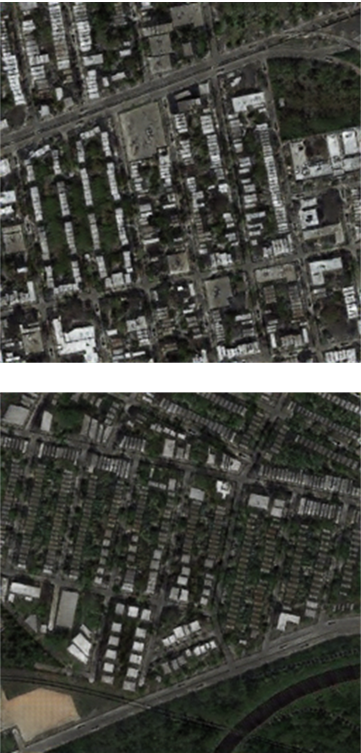}&
    \includegraphics[width=.168\linewidth]{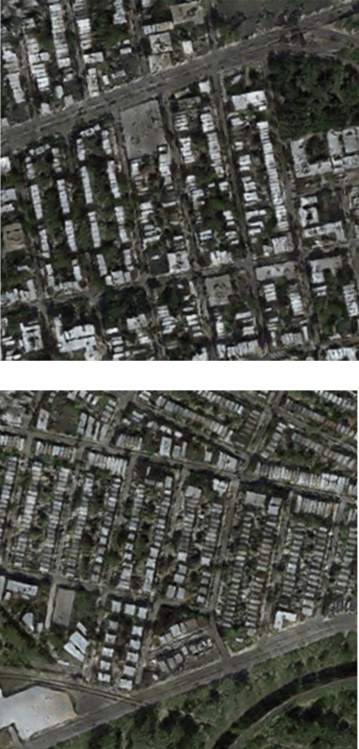}&
    \includegraphics[width=.168\linewidth]{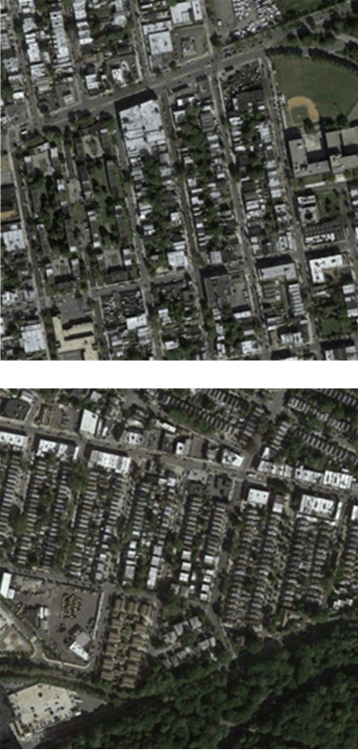}    \\
    (a) & (b) & (c) &(d) &(e)\\
    \end{tabular}
  \caption{CycleGAN results for mapping (a) real maps to fake images in the aerial photos domain for (b) unbalanced Adam, (c) unbalanced MTAdam and (d) balanced Adam. (e) is the ground truth.}
    \label{fig:cyclegan-quali}
\end{figure*}

Fig.~\ref{fig:cyclegan-quali} exhibits representative images from the CycleGAN models, showing that MTAdam, even when applied to a loss with unbalanced weights, matches the results of Adams on the prescribed weights.

{\bf SRGAN\quad} In the Super Resolution GAN (SRGAN)~\cite{ledig2017photo}, the objective function is:
\begin{equation}
\pazocal{L}_{SRGAN} =  \lambda_1 \pazocal{L}_{MSE} + \lambda_2 \pazocal{L}_{GAN} + \lambda_3 \pazocal{L}_{Perceptual} 
\end{equation}
for which the total loss is a combination of a GAN loss, perceptual loss and MSE. The $\lambda_1$, $\lambda_2$ and $\lambda_3$ are set to 1, 0.001, 0.006. We unbalance SRGAN experiments by setting both $\lambda_2$ and $\lambda_3$ to 1 (i.e. applying all three terms with the same coefficient). 

We evaluate SRGAN-Unbalanced-MTAdam on two test sets, Set14 \cite{set14} and BSD100\cite{BSD100}. The results, listed in Tab.~\ref{Tab:SRGAN_test} demonstrate that MTAdam can effectively recover from unbalanced weights, while the other optimization methods suffer a degradation in performance.

\begin{table*}[t!]
\caption{SRGAN results. Higher is better for PSNR and SSIM. Lower is better for FID. }
\label{Tab:SRGAN_test}
\centering
\begin{tabular}{lc@{~~}c@{~~}cc@{~~}c@{~~}c}
\toprule
&\multicolumn{3}{c}{Set14} & \multicolumn{3}{c}{BSD100}\\
\cmidrule(lr){2-4}\cmidrule(lr){5-7}
&  \textbf{PNSR} &  \textbf{SSIM} & \textbf{FID} & \textbf{PSNR}& \textbf{SSIM} &  \textbf{FID} \\
\midrule
Unbalanced-SGD-Momentum & 22.95 &  0.6213 & 74.92  &  23.01 & 0.5997 & 79.45  \\
Unbalanced-RMSProp & 23.04 & 0.6252 & 72.92  &  23.14& 0.6020 & 76.64  \\
Unbalanced-Adam & 23.43 &   0.6264 & 69.96  &  23.71 & 0.6070 & 77.24  \\
Unbalanced-MTAdam & 25.97 & 0.7219 & {\textbf{66.03}} & \textbf{25.23}& {\textbf{0.6690}} & {74.97} \\
\midrule
Hyperband-FID
& 24.89 & 66.24 & 67.26 & 24.80 & 0.6554 & 75.65\\
\midrule
Balanced-Adam & \textbf{26.02} & \textbf{0.7397}  & 66.59& 25.16 & 0.6688& \textbf{72.57} \\
\bottomrule
\end{tabular}
%\end{table*}
%\begin{table*}[t!]
~\\
\smallskip
\caption{Ablation study results.}
\label{Tab:ablation}
\centering
\begin{tabular}{@{}l@{~}l@{~}c@{~}c@{~}c@{~}c@{~}c@{~}c@{~}c@{~}c@{}}
\toprule
&& \multicolumn{4}{c}{pix2pix facade} & \multicolumn{4}{c}{CycleGAN horse2zebra}\\
\cmidrule(lr){3-6}
\cmidrule(lr){7-10}
& &  \textbf{L1} &  \textbf{L2} &
\textbf{NMSE} & \textbf{FID}&  \textbf{L1-A} &  \textbf{FID-A} &
\textbf{L1-B} & \textbf{FID-B}  \\
\midrule
% &Unbalanced-Adam &  \\
(i)& Removing L6--8 & 53.45 & 5021 & 0.326  &  134.0 &74.7&225.2&71.2 & 217.1\\
(ii)& Removing L6, changing $g^i_{\ell}$ to $g^i$ & 46.8 & 3850 & 0.258  & 133.5 & 70.1 & 75.2 & 69.5 & 145.7\\
(iii)& No scaling by $n_{\ell,t}^1$ in L8 & 46.7 & 3862 & 0.242  &  136.3 & 72.3 &  82.4& 70.6& 167.2\\
(iv)& Line 14: Scaling like Adam& 48.2 & 3956 & 0.245  &  171.5 & 73.5 & 210.5 & 71.1& 190.3\\
(v)&Line 14: scaling by mean instead of max & 48.3 & 4020 & 0.249 & 156.8 & 73.7& 208.4& 71.6&173.5\\
& Full method & \textbf{46.5} & \textbf{3822} & \textbf{0.236} & \textbf{130.4} &  \textbf{68.7} & \textbf{73.0} & \textbf{69.0}  & \textbf{139.7}\\
\bottomrule
\end{tabular}
\end{table*}

\subsection{Ablation Study}
\label{ablation}

Tab.~\ref{Tab:ablation} presents an ablation study for unbalanced pix2pix on the facade images and for unbalanced CycleGAN on the zebra2horse dataset. The following variants are considered (the descriptions refer to Alg.~\ref{algo:MTAdam}): (i) treating all layers as one layer and eliminating lines 6-8 altogether. (ii) training all layers at one, and performing the normalization in line 8 once for the entire gradient, i.e., still normalizing by the magnitude of the gradient of the first term. (iii) scaling the gradients of each layer $l$ and each term $i$ in line 8 by $(n_{l,t}^i)^{-1}$ but not by $n_{l,t}^1$. (iv) replacing the term ${\max(\widehat{v^1_t}...\widehat{v^I_t})}$ in line 14 with ${\widehat{v^i_t}}$, in an analog way to line 14 in Adam. (v) replacing the same term with the mean $I^{-1}\sum \widehat{v^i_t}$.

The results, shown in Tab.~\ref{Tab:ablation}, indicate that it is crucial to employ a per layer analysis, in the way it is done in MTAdam, that normalizing by the magnitude of the gradient of an anchor term is highly beneficial, and that the maximal variance is a better alternative to alternative scaling terms.

\section{Conclusions}

MTAdam is shown to be a widely applicable optimizer, which can dynamically balance multiple loss terms in an effective way. While tested on image generation tasks, it is a general algorithm, which can find its usage in other types of tasks that require the optimization of multiple terms, such as domain adaptation and some forms of self-supervised learning. Our code can be found here\footnote{https://github.com/ItzikMalkiel/MTAdam}. MTAdam is implemented as a generic pytorch optimizer and applying it is almost as simple as applying Adam.

\section*{Acknowledgements}
This project has received funding from the European Research Council (ERC) under the European Unions Horizon 2020 research and innovation program (grant ERC CoG 725974).

{\small
\bibliographystyle{plain}
\bibliography{Adam.bib}

}

\clearpage
\begin{center}
{\LARGE Appendix}
\end{center}
\appendix

\section{Image synthesis samples}

\begin{figure*}[t]
  \includegraphics[width=1.0\linewidth]{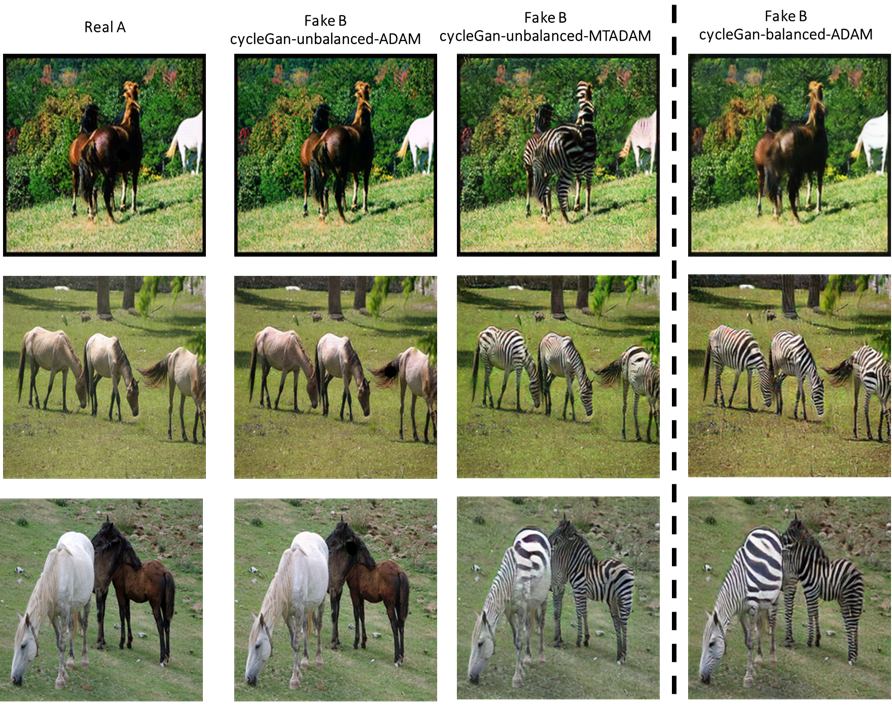}
  \centering
  \caption{Samples from the CycleGAN experiments employed on the zebra2horse dataset, mapping horse images to zebras. CycleGAN-Balanced-Adam employs CycleGAN with the prescribed weights~\cite{zhu2017unpaired}. CycleGAN-Unbalanced-Adam runs with unbalanced initial weights, which fails to generate zebra images.  CycleGAN-Unbalanced-MTAdam (our method) starts with unbalanced weights but produces images of the same quality as  CycleGAN-Balanced-Adam.
  }
    \label{fig:cyclegan-quali2}
\end{figure*}

Fig.~\ref{fig:cyclegan-quali2} presents a few representative samples from the CycleGAN\cite{zhu2017unpaired} experiments, employed to transfer horse images to zebras. As can be seen, the unbalanced Adam training completely fails to generate zebras images and collapses to the identity mapping. This can be attributed to the domination of cyclic loss in the unbalanced settings, which dictates the convergence, leaveing the other loss terms ineffective. On the other hand, the unbalanced MTAdam model was able to successfully generate zebra images, in a quality that is similar or better to the balanced adam model (which uses the prescribed weights). In particular, in the first row, we can see that CycleGAN-Unbalanced-MTAdam was able to outperform the CycleGAN-Balanced-Adam experiment, as the latter fails to generate a zebra image for this particular sample, while MTAdam was able to generate a fairly good quality image.

\begin{figure*}[t]
  \includegraphics[width=1.0\linewidth]{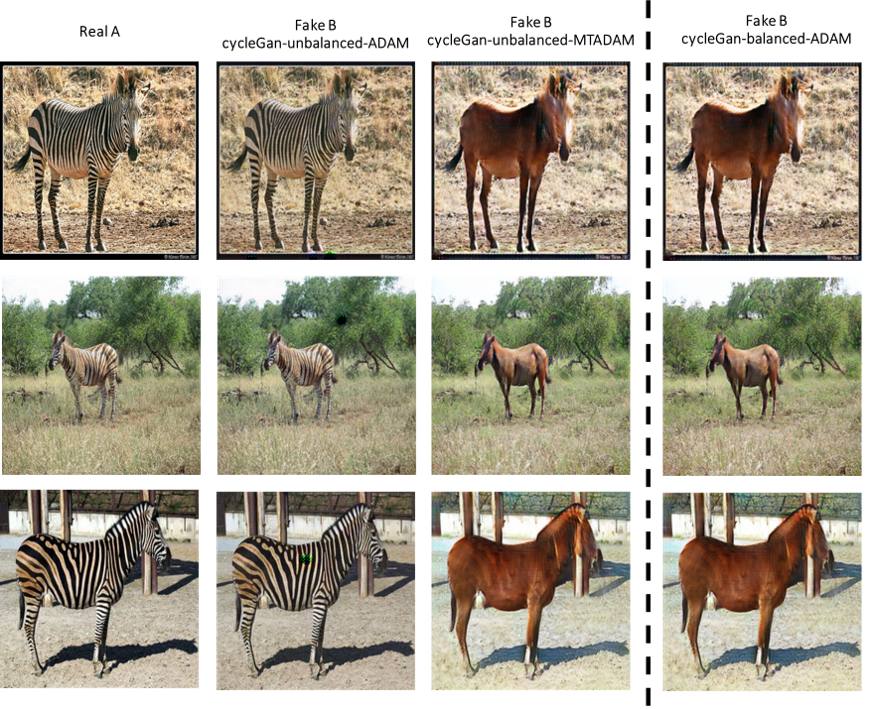}
  \centering
  \caption{Samples from the CycleGAN experiments employed on the zebra2horse dataset, mapping zebra images to horses. CycleGAN-Balanced-Adam employs CycleGAN with the prescribed weights~\cite{zhu2017unpaired}. CycleGAN-Unbalanced-Adam runs with unbalanced initial weights, fails to generate horse images, collapses to identity mapping, and introduces visual artifacts.  CycleGAN-Unbalanced-MTAdam (our method) starts with unbalanced weights but produces images of the same quality as  CycleGAN-Balanced-Adam.
  }
    \label{fig:cyclegan-quali3}
\end{figure*}

Fig.~\ref{fig:cyclegan-quali3} depicts a few samples from the same models described above, this time employed to generate horse images from zebra images. As can be seen, the unbalanced Adam fails again to generate horse images, collapses to the identity mapping, and this time also introduces visual artifacts in a few images (see the green artifact in the bottom row). In addition, MTAdam yields images of the same quality as the balanced Adam experiment. 

\begin{figure*}[t]
  \includegraphics[width=1.0\linewidth]{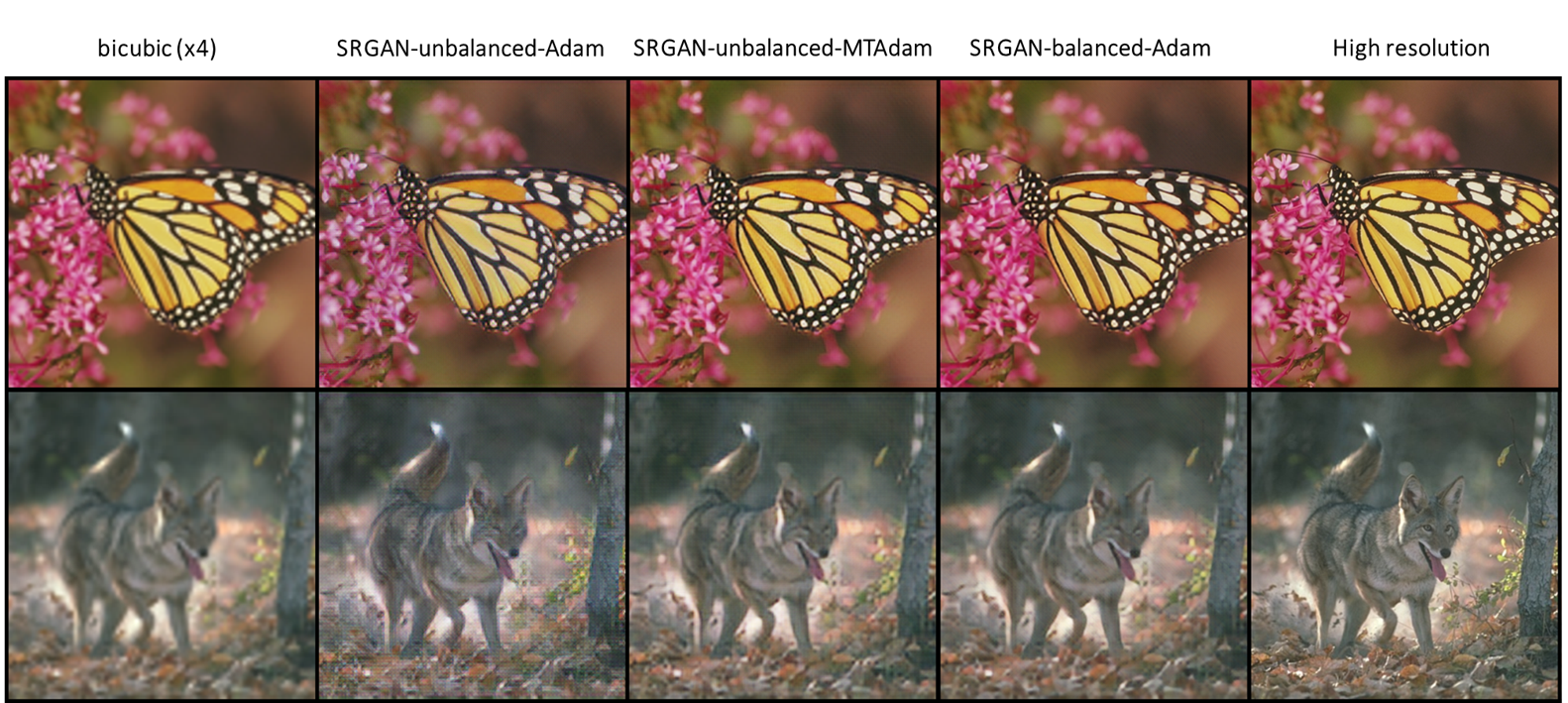}
  \centering
  \caption{Samples from the test set of the SRGAN\cite{ledig2017photo} experiments. Images were taken from Set14\cite{set14}. SRGAN-Unbalaced-Adam employs SRGAN training with the unbalanced initial weights, as described in the main text. SRGAN-Balanced-Adam employs SRGAN with the prescribed weights. SRGAN-Unbalanced-MTAdam utilizes our proposed MTAdam optimizer, runs with unbalanced initial weights but produces images of the same quality of SRGAN-Balanced-Adam. 
  }
    \label{fig:SRGAN-quali3}
\end{figure*}

Fig.~\ref{fig:SRGAN-quali3} presents visual results for the SRGAN \cite{ledig2017photo} experiments. The images were taken from Set14 \cite{set14}. All models were trained to generate high-resolution images from low-resolution images, with a factor of 4x upscaling. As can be seen, the SRGAN model that employs unbalanced weights and Adam optimizer yields images with visual artifacts, and low fidelity. This can be attributed to the domination of the GAN and perceptual loss terms over the pixel-wise term. In contrast, the SRGAN model that employs our MTAdam optimizer with unbalanced weights yields images of the same quality as the SRGAN that utilizes the prescribed weights \cite{ledig2017photo} and Adam.

\end{document}